\documentclass[letterpaper, 10pt, conference]{ieeeconf}
\usepackage{times}

\IEEEoverridecommandlockouts                              

\usepackage{mathtools}
\usepackage{makeidx}   
\usepackage{graphicx}  
        
\graphicspath{ {./} {./fig/} }

\usepackage{multicol}  
\usepackage[bottom]{footmisc}
\usepackage{subcaption}
\usepackage[dvipsnames,svgnames]{xcolor}

\usepackage{amsmath,amssymb,amsthm}
\usepackage{calc}
\usepackage{tikz}
\usepackage{xfrac}
\usepackage{array}
\usepackage{bm}
\usepackage[utf8]{inputenc}

\newtheorem{problem}{Problem}

\newif\ifmargincomments 
\margincommentsfalse

\newif\ifjournalv 
\journalvfalse

\ifmargincomments

\newcommand{\frmargin}[2]{{\color{Green}#1}\marginpar{\color{Green}\raggedright\footnotesize [FR]:#2}}

\else
\newcommand{\frmargin}[2]{#1}

\fi

\newcommand{\mdpaction}{\ensuremath{a}}

\newif\ifrelaxedv  
\relaxedvfalse
\ifrelaxedv
\newcommand{\jvspace}[1]{}
 \else
\newcommand{\jvspace}[1]{\vspace{#1}}
\renewcommand{\baselinestretch}{.942}
\fi

\ifjournalv
\newcommand{\journal}[1]{#1}
\else
\newcommand{\journal}[1]{}
\fi

\title{\LARGE \bf{Stochastic Guidance of Buoyancy Controlled Vehicles under Ice Shelves using Ocean Currents}}
\author{Federico Rossi$^1$, Andrew Branch$^1$, Michael P. Schodlok$^1$, Timothy Stanton$^2$, \\ Ian G. Fenty$^1$, Joshua Vander Hook$^1$, Evan B. Clark$^1$
\thanks{$^1$ Jet Propulsion Laboratory, California Institute of Technology, Pasadena, CA, 91109; {\tt \{federico.rossi, andrew.branch, michael.p.schodlok, ian.fenty, hook evan.clark\}@jpl.nasa.gov}.}
\thanks{$^2$ Moss Landing Marine Laboratories; {\tt stanton@nps.edu}}
}

\begin{document}
 \bstctlcite{IEEEexample:BSTcontrol}

\maketitle

\begin{abstract}
We propose a novel technique for guidance of buoyancy-controlled vehicles in uncertain under-ice ocean flows.
In-situ melt rate measurements collected at the grounding zone of Antarctic ice shelves, where the ice shelf meets the underlying bedrock, are essential to constrain models of future sea level rise.
Buoyancy-controlled vehicles, which control their vertical position in the water column through internal actuation but have no means of horizontal propulsion, offer an affordable and reliable platform for such in-situ data collection. 
However, reaching the grounding zone requires vehicles to traverse tens of kilometers under the ice shelf, with approximate position knowledge and no means of communication, in highly variable and uncertain ocean currents. To address this challenge, we propose a partially observable MDP approach that exploits model-based knowledge of the under-ice currents and, critically, of their uncertainty, to synthesize effective guidance policies. The approach uses approximate dynamic programming to model uncertainty in the currents, and QMDP to address localization uncertainty.
Numerical experiments show that the policy can deliver up to 88.8\% of underwater vehicles to the grounding zone -- a 33\% improvement compared to state-of-the-art guidance techniques, and a 262\% improvement over uncontrolled drifters. Collectively, these results show that model-based under-ice guidance is a highly promising technique for exploration of under-ice cavities, and has the potential to enable cost-effective and scalable access to these challenging and rarely observed environments.
\end{abstract}

\section{Introduction}\label{sec:introduction}

Ice shelves are vast, floating slabs of ice that fringe 75 percent of the Antarctic coastline and act as ``corks in the bottle'', preventing the rest of the ice on the continent from sliding into the ocean and catastrophically raising global sea levels. By the end of the century, the collapse of Antarctic ice shelves could trigger a meter or more of sea level rise, with profound effects for hundreds of millions of people worldwide. In total, Antarctic ice shelves hold back enough ice to raise global sea levels by more than fifty meters \cite{Harig2015-uk}. Yet a lack of detailed understanding about how ice shelves will behave in a warming climate remains a primary obstacle to accurate projections of sea level rise.

Current state-of-the-art sea-level rise projections have extremely large uncertainties. Specifically, the latest IPCC special report states that, by the end of the century, sea-level rise could range between $0.29$ m and $1.1$ m, depending on emission scenarios, associated climate policy, and the response of the Antarctic Ice Sheet as the world continues to warm \cite{Oppenheimer2019-xa}. Some studies suggest that sea-level rise of as much as two meters is possible by 2100 \cite{Church2013-cr}.

The single largest drivers of sea-level rise uncertainty are poorly-constrained numerical models of ice shelf melt and collapse, which suffer from a dearth of in-situ measurements to provide ground-truth for basal melt rates under ice shelves. Although various assets, ranging from underwater vehicles to borehole-deployed instruments, have managed to collect some in-situ data sets beneath ice shelves, these data sets usually do not observe basal melt rate, and are also severely limited in duration, location, and spatial distribution. Basal melt rates can be estimated from remote sensing by examining the residual differences between surface elevation change data from satellite altimeters and dynamical volume convergence calculations from combined ice field velocity, surface snow/ice mass flux, and firn compaction estimates \cite{adusumilli2020interannual}. However, large uncertainties remain due to unknown local atmospheric properties and firn behavior; critically, such techniques are inadequate at large ice-shelf grounding zones, which are primary contributors to melt, due to severe surface crevassing which does not allow for easy estimation of the aforementioned effects.


\begin{figure}[t]
\jvspace{-1.15em}
\centering
\includegraphics[width=.5\textwidth]{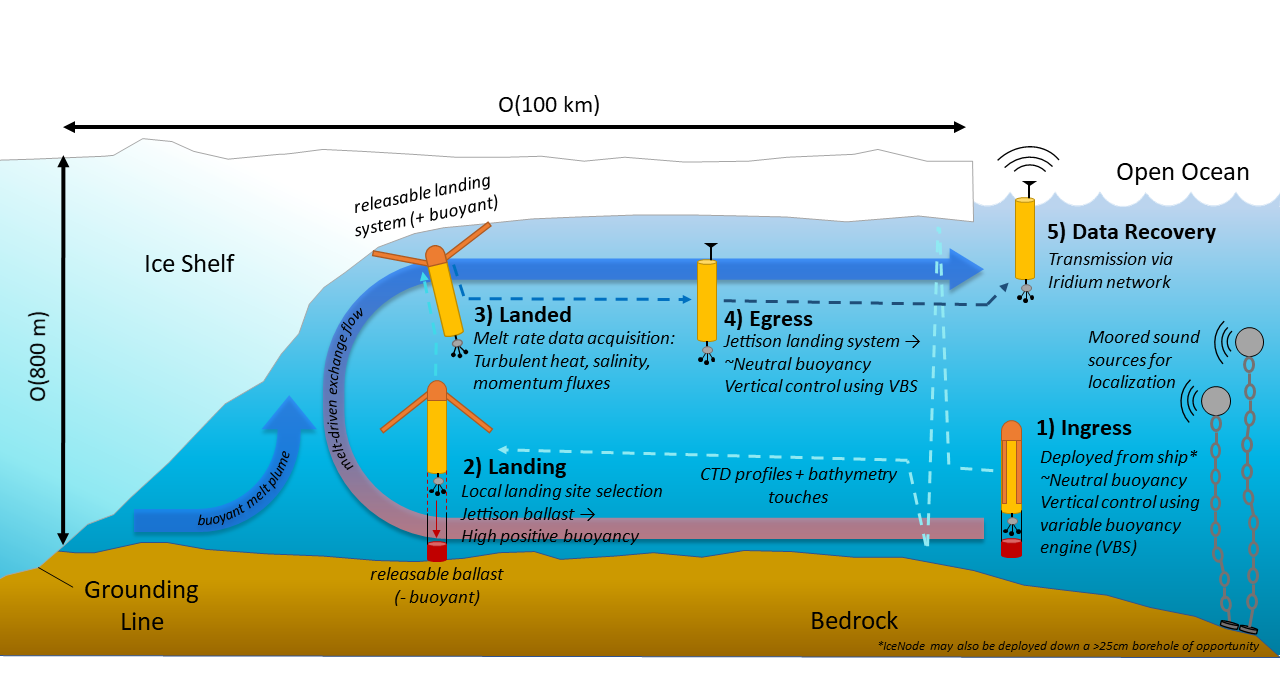}
\caption{IceNode concept of operations and mission phases}
\label{fig:conops}
\end{figure}

The grounding zone of an ice shelf is the area where the shelf first becomes buoyantly decoupled from the bedrock below it. Since the grounding zone is where ice first comes in contact with the warm, high-salinity, dense water entering at depth under the ice shelf, it is critical to capture the ice-ocean interactions as close as practical to this region. Due to large strain gradients created by the ice shelf ``hinge" formed at this transition, grounding zones are typically heavily crevassed, making melt rate estimations from remote sensing difficult. While very limited melt rate and boundary layer turbulent fluxes have been measured under selected Antarctic ice shelves using borehole-deployed instruments \cite{Stanton2013-pj} \cite{Davis2020-np}, no measurements have been made in active, warm ice shelves in close proximity to the grounding zone, as crevassing prevents borehole drilling operations. Furthermore, borehole based deployments are logistically complicated, expensive, and difficult to scale across multiple measurement sites concurrently for comparative studies. 

In line with this, a recent Keck Institute for Space Studies workshop, \textit{The Sleeping Giant: Measuring Ocean-Ice Interactions in Antarctica} \cite{Thompson2015-uu}, identified long-duration measurements of melt rate as critically important to understand grounding zone dynamics and future sea-level rise, and recommended development of autonomous guidance capabilities to enable under ice vehicles to conduct cost-effective, practical, and persistent monitoring, returning long-duration ground-truth datasets from near the grounding zone.

In this paper we describe and demonstrate in simulation an autonomous guidance technique for a novel underwater vehicle, IceNode, designed to drift underneath ice shelves on melt-driven exchange currents, using buoyancy control but no propulsion, and then land against the underside of the ice to directly measure basal melt rate. 

Our technique enables IceNode vehicles, deployed at the shelf edge by an ice breaker, to navigate to regions near the grounding zone to conduct landed science, and then return to open ocean to relay the collected data to scientists. Although the proposed technique is designed and evaluated for the IceNode vehicle, the approach is generally applicable to under-ice buoyancy-controlled vehicles navigating in melt-driven exchange flows, and has the potential to enable reliable and cost-effective access to these challenging and rarely observed environments.

\subsection{IceNode: System Description}
\label{sec:system}

IceNode is a buoyancy-driven vehicle with similar functionality to vertically profiling floats (e.g. \cite{Sanford2005-nf}), but specifically designed to gather in-situ melt rate observations at the basal melt interface of large, difficult to access ice shelves. IceNode
controls
 its vertical position in the water column using a variable buoyancy system (VBS) which pumps oil between an external bladder and the internal pressure hull to change density, and thus sink deeper or float higher. Compared to vertically profiling floats, \journal{which operate only in the water column, }IceNode offers the additional capability to buoyantly land against the underside of the ice shelf to acquire stable fixed measurements of turbulent heat, salt, and momentum fluxes using direct eddy-correlation techniques, which can be used to calculate in-situ melt rate \cite{Stanton2013-pj}. Figure \ref{fig:conops} shows the distinct stages of an IceNode mission. First, IceNode is deployed in open ocean near the shelf edge by an ice breaker. Next, in the ingress phase, IceNode exploits currents at different depths to navigate to a pre-specified target area near the grounding zone (this technique is the subject of this paper). While drifting in the water column, IceNode is localized using acoustic multilateration from moored sound sources placed at the shelf edge, using the same technique successfully demonstrated with EM-APEX floats in \cite{Girton2019-js}. Once beneath the target area, IceNode ascends to a fixed standoff from the ice and uses an upward-looking Doppler Velocity Log (DVL) to locate a suitable landing location by examining surface slope and roughness. \frmargin{Once an appropriate landing location is found}{Could shorten}, IceNode deploys landing legs, releases a ballast weight to become highly positively buoyant, and lands against the underside of the ice. The vehicle then collects in-situ measurements of heat, salt, and momentum fluxes at the basal melt interface for a year. Once the landed phase is complete, IceNode jettisons its highly positively buoyant landing legs to achieve near-neutral buoyancy again and exploits melt-driven exchange currents to egress back to open water. Finally, the vehicle surfaces and transmits its mission data back to scientists over an Iridium link (the vehicle is not physically recovered). IceNodes are designed to be cheap (by underwater vehicle standards) and expendable, and multiple IceNodes are concurrently deployed at the shelf edge by an ice breaker. Individual vehicles can be directed to land in different regions under the shelf, and thus form an array of instrument platforms that acquires long-duration, concurrent, well-distributed time series of basal melt rate. The capability of IceNode to land and acquire direct melt rate measurements at the basal melt interface is unique among underwater vehicles, and the drift-based access technique enables long mission duration and cost-effective targeting of areas near the grounding zone not achievable with traditional borehole-deployed instrument packages.

\subsection{State of the Art}

The cavities underneath ice shelves are notoriously difficult to access and return safely from, and are cut off from communication with the outside world by up to a thousand meters of ice overhead. \journal{For this reason, in-situ data sets from these environments are relatively few and far between. }IceNode's concept of operations draws heavy inspiration from the successful 2019 University of Washington Applied Physics Lab (APL-UW) campaign of four EM-APEX floats beneath the Dotson Ice Shelf \cite{Girton2019-js}. This campaign depended on melt-driven exchange flow to move the vehicles into, around, and out of the cavity. After deployment at the shelf edge by an ice breaker, the EM-APEX floats used a VBS to descend to a depth where they were swept underneath the cavity by deep inflow currents. During the ingress phase, the floats maintained a depth corresponding to 75\% of the water column depth, periodically computed by bumping against the seafloor and the ice shelf base. After a pre-set timer elapsed, the floats transitioned to egress, and moved to 25\% of the cavity depth to be swept out to sea on shallow outflowing currents.
Throughout the mission, the floats collected conductivity, temperature, pressure, and current data, as well as recorded ranging signals from a set of three RAFOS acoustic moorings placed at the shelf edge. Using this technique, all four EM-APEX floats eventually emerged from the cavity, after spending multiple months and collectively traveling hundreds of km under the shelf, demonstrating that riding melt-driven exchange driven flows from the shelf edge is a viable technique for exploring ice shelf cavities.

Other successful missions have been conducted using AUVs \cite{McPhail2009-ee} \cite{Davies2017-xl} \cite{McPhail2019-ec} and gliders \cite{Lee2018-yk} from the shelf edge, and cabled instrumentation \cite{Stanton2013-pj} \cite{Davis2020-np} and HROVS deployed through boreholes \cite{Lawrence2020-dn} \cite{Schmidt2020-xt} . However, with the exception of long-duration borehole-deployed instruments and the APL-UW Dotson gliders and floats, these missions are typically short-lived (on the order of hours to days), only deploy a single vehicle or asset, and none directly provide long duration, spatially-distributed concurrent melt rate data sets directly at the basal melt interface.

Much research exists related to path planning of under-actuated marine vehicles in flow fields using ocean circulation models, including efficient long-range path planning of gliders in the presence of currents \cite{Rao2009-hw} \cite{Thompson2010-ny}, stationkeeping of gliders \cite{Clark2019-yh} and vertically profiling floats \cite{Troesch2018-hm} near a location of interest, avoiding glider surfacing in dangerous locations \cite{Pereira2013-uw}, and optimizing float coverage across oceans \cite{Dahl2011-lb}.  Similar techniques exist for path planning of aerostats in wind fields, including Google's Project Loon using superpressure stratospheric balloons to provide internet connectivity \cite{Bellemare2020-cr}, and planetary mission concepts for future Venus \cite{Wolf2010Venus} and Titan \cite{Fathpour2014Venusgraph} missions. 
The majority of these works generally assume that the the circulation model is \emph{known} (either numerically or through accurate measurements), or that updated model predictions based on external measurements can be periodically communicated to the vehicle, enabling the use of deterministic path planning algorithms - reasonable assumptions for atmospheric and surface navigation, but unrealistic ones for the communication-denied environment considered in this paper. The work in
 \cite{Wolf2010Venus} does use a stochastic model of the flow field, but it employs an extremely simple probabilistic model to capture the variability of currents.
 

\subsection{Contribution}
Our contribution is twofold.
First, we propose a novel approach for model-based under-ice guidance under model uncertainty. The approach, based on approximate dynamic programming, exploits model information to compute policies that exploit the currents for guidance with only vertical actuation; critically, it accommodates \emph{model uncertainty}, rather than relying on a (possibly inaccurate or outdated) deterministic representation of under-ice currents. The approach is heavily inspired by \cite{Wolf2010Venus};  we extend this work by (i) proposing a rigorous and systematic way of capturing the flow distribution from time-varying model data, and (ii) accommodating position uncertainty.  

Second, we validate the approach through extensive numerical experiments set beneath the Pine Island Glacier ice shelf in Antarctica. Simulation results show that the proposed approach can can deliver up to 88.8\% of underwater vehicles to the grounding zone -- a 33\% improvement compared to state-of-the-art under-ice guidance techniques for buoyant vehicles, and a 262\% improvement over uncontrolled vehicles. The fraction of vehicles that reaches the grounding zone can be further increased up to 95\% as localization uncertainty is reduced. Collectively, these results show that model-based under-ice guidance holds promise to enable previously-infeasible measurements of melt rates at the grounding zone of Antarctic glaciers in a cost-effective manner, providing critical in-situ measurements to improve sea-level rise models, and informing climate science and public policy.

\subsection{Organization}
The rest of this paper is organized as follows. In Section \ref{sec:problem-statement} we formally state the under-ice guidance problem we wish to solve, and discuss assumptions. In Section \ref{sec:cavity-model} we describe the numerical model used to characterize under-ice flows, and discuss its relevance to the planning problem. Section \ref{sec:navigation} presents the proposed approach to model-based under-ice guidance. The effectiveness of the proposed approach is assessed through numerical simulations in Section \ref{sec:experiments}. Finally, in Section \ref{sec:conclusions}, we present our conclusions and lay out directions for future research.
        
\section{Problem Statement}
\label{sec:problem-statement}

The goal of this paper is to provide an efficient algorithm for autonomous under-ice guidance of buoyancy-controlled vehicles in a partially-unknown flow field.

We assume that a \emph{stochastic} model of the flow field in a region of interest is available. For each point in the domain, the model specifies the probabilistic distribution of the flow field that may be encountered at that location. Such a model can be obtained through, e.g., numerical simulations with state-of-the-art circulation models, \frmargin{numerically capturing uncertainty due to the initial conditions and assessing its impact on cavity flow uncertainty.}{Ambiguous}

An autonomous vehicle navigates in the flow field. The vehicle can control its vertical location in the water column using a variable-buoyancy mechanism. The vehicle has no other means of propulsion, and its horizontal position evolves according to the flow field as a semi-Lagrangian tracer. 
The vehicle has access to a probabilistic distribution (or \emph{belief}) of its likely location in the flow.
A number of regions in the flow field are designated as \emph{end regions}; each region is associated with a reward, which captures the scientific interest of reaching that region.
The vehicle expends energy to control its position; we assume that the energy expenditure has a constant component which characterizes the ``hotel load'' for non-propulsion functions, and a variable component which captures the energy cost to ascend in the water column.


We are now in a position to formalize the problem that we wish to solve. 

\begin{problem}[Autonomous buoyancy-controlled guidance in uncertain flow field]
\label{problem:navigation}
Given a stochastic model of a flow field, a set of end regions of interest, and costs capturing the vehicle's energy expenditure, compute an optimal policy (i.e. a mapping from beliefs about the vehicle location to desired controlled depths) that maximizes the total discounted reward obtained by the vehicle, i.e. the expected discounted reward for reaching a region of interest minus the expected discounted  energy cost incurred along the trajectory.
\end{problem}


\section{Ice Shelf Cavity Modeling}
\label{sec:cavity-model}

Availability of high-quality stochastic models of under-ice circulation is critical to the proposed guidance technique. 

\begin{figure}[h]
\centering
\includegraphics[width=.47\textwidth]{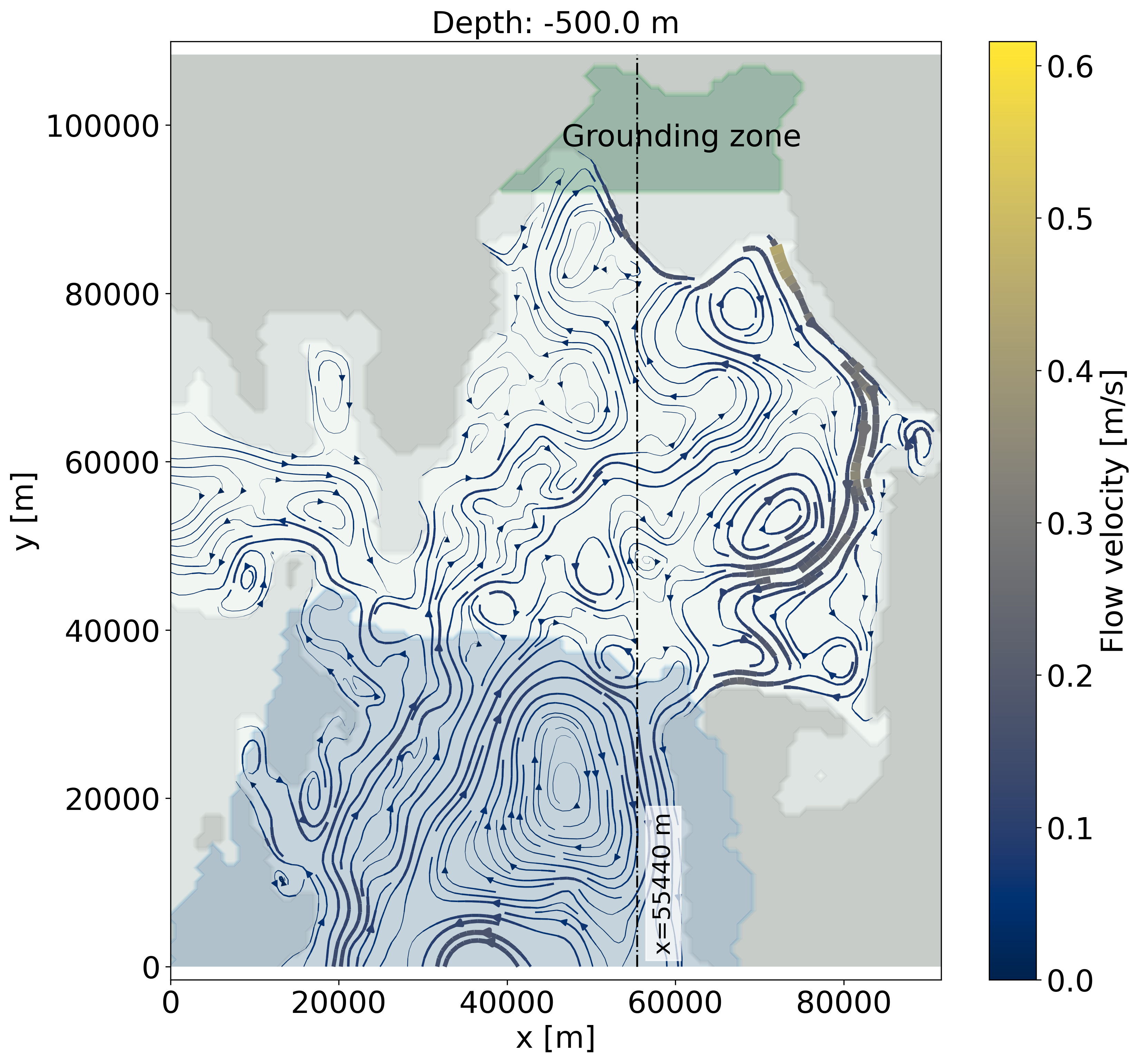}
\includegraphics[width=.45\textwidth]{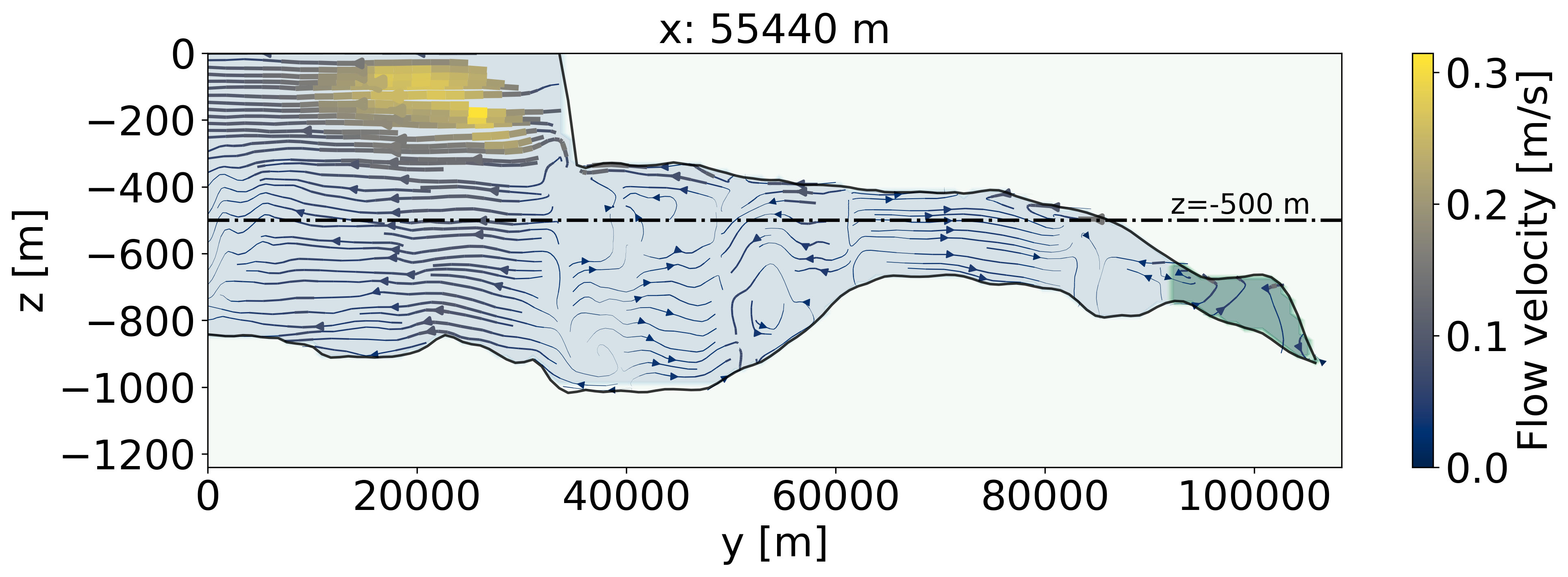}
\caption{Instantaneous currents under the Pine Island ice shelf at $500$ m depth and across one vertical slice for one time step.} 
\label{fig:streamlines} 
\end{figure}

In this paper, we simulate ice-ocean interaction using the Massachusetts Institute of Technology general circulation model (MITgcm), which includes a dynamic/thermodynamic sea-ice model \cite{losch10} and captures the temporal evolution of under-ice currents, temperature, and salinity. Freezing and melting processes in the sub-ice-shelf cavity are represented by the three-equation thermodynamics of Hellmer and Olbers  
\cite{hellmer89} with modifications by Jenkins  \cite{jenkins01}. 
The model domain, the Pine Island Ice Shelf (shown in Figure \ref{fig:streamlines}), is derived from the global cube-sphere configuration (CS510) used by  the ECCO2 project  \cite{Menemenlis2008ECCO2}, with a nominal horizontal grid spacing of $280$ m and 250 vertical levels, each with $5$ m thickness. The bathymetry and ice shelf draft are provided by BedMachine Antarctica \cite{morlighem20}. Initial conditions and boundary conditions for hydrography (temperature, salinity, and horizontal velocity components $u$ and $v$) and sea ice (concentration, ice thickness, and snow thickness) are derived from a global, coarser resolution ($\sim 20$ km horizontal grid spacing), data-constrained solution for the period of 2009--2012. Due to the difference in resolution between the global model and the $280$ m model domain, a relaxation (10 grid points into the model domain) is applied to temperature and salinity at the boundaries to avoid artifacts such as wave energy radiating into the model interior; similarly, a 5-grid-point relaxation is used for sea ice variables. Surface forcing is provided by the ERA-Interim reanalysis project \cite{dee11}. Similar configurations have been successfully applied to study ice shelf ocean interaction on the cube sphere grid (e.g., \cite{schodlok12}, \cite{schodlok16}, \cite{khazendar13}). 

\frmargin{
The model uses an Arakawa C-Grid, where the three current velocity components are not co-located in a grid cell \cite{arakawa1977}.
Four-dimensional $(x,y,z,\text{time})$ interpolation is applied for each velocity component independently.
To determine the portion of the model containing navigable water, the portion of each cell that contains water (as opposed to land, or ice) is computed.
The vertical depth boundaries of the model are determined using this data from the center of each cell, linearly interpolated in the $x$ and $y$ dimensions.
The horizontal bounds of the navigable portion of the model are determined as any location where $4$D gridded depth interpolation is possible. This functionally truncates the bounds of the model by half a grid cell ($\sim 140$ m). 
These limitation are acceptable as the affected areas are below the resolution of the model, and only along the boundaries of the model, where it is undesirable for vehicles to travel.}{This can be streamlined a bit}

Figure \ref{fig:streamlines} shows a snapshot of the model output (specifically, the cavity flows) for one time step.

The simulated flow is consistent with the geophysical fluid dynamical and ice melt equations encoded in the numerical model, with the prescribed seafloor and ice-shelf cavity geometry, and with external atmospheric and open-boundary forcing. While significant uncertainties remain in the ice shelf cavity geometry and in atmospheric and open boundary forcing, we have confidence in many aspects of the simulated ice shelf cavity circulation, such as the propensity for the balanced flow to follow contours to conserve potential vorticity ($f/h$, where $h$ is the floor-ceiling column thickness), and the cavity-scale overturning circulation in which relatively dense, salty warm waters flow towards the grounding line at depth, and return towards the cavity entrance near the cavity ceiling as relatively lighter fresh cool waters. Of course, specific aspects of the time-variable, fine-scale turbulent circulation (i.e., the instantaneous arrangement of meso- and submeso-scale eddies and filaments) are assumed to be realistic and representative in a \emph{statistical} sense. Therefore, while the true distribution of velocities within the cavity is unknown (and will remain so for all practical purposes),  the numerical model provides a reasonable and useful approximation for the flow distribution that a float would encounter. 


\section{Guidance in uncertain flow fields}
\label{sec:navigation}

We are now in a position to describe the proposed approach to solve Problem \ref{problem:navigation}.
First, we present a continuous-space MDP formulation that leverages the MITgcm flow solution, which we solve through approximate dynamic programming (ADP) \cite{bertsekasdynamicprogramming2012} to compute an optimal policy when the vehicle's position is exactly known. Next, we discuss how QMDP \cite{littman1995learning} can be used to extend the applicability of the policy to the case where the vehicle's position is only approximately known.

\subsection{Continuous-space Markov Decision Process}
\label{sec:navigation:mdp}

We formalize the under-ice guidance problem as a continuous-state Markov Decision Process by defining its states, actions, transitions, rewards, and final states. We discretize time according to a discrete time step $\delta$.

\paragraph{States} The state of the vehicle is the vehicle's location under the ice. Formally, the set of states is:
\[
\mathcal{S} = \{(x,y,z) | (x,y,z)\in \text{navigable water} \cap z>\underline{z}\},
\]
where the navigable region under the ice is computed according to the model described in Section \ref{sec:cavity-model}, and the maximum allowable depth $\underline{z}$ captures the vehicle's depth rating.

\paragraph{Actions} The vehicle can choose to move to a different depth $\mdpaction$ through a buoyancy control mechanism. 
The vehicle's ascent and descent rate are constrained to be lower than a given maximum and minimum rate
$\overline{\dot z}$ and $\underline{\dot{z}}$ respectively, 
and the vehicle should ensure that the desired depth will be within navigable waters. Formally, the actions available in state s are the set of depths:
\[
\mathcal{A}((x,y,z)) = \{\mdpaction | \delta \underline{\dot{z}} \leq (\mdpaction-z)< \delta \overline{\dot{z}}  \cap (x,y,\mdpaction) \in \mathcal{S}\}
\]

\paragraph{Final States}
Certain states $\mathcal{F}\subset\mathcal{S}$ are denoted as final states: when the vehicle reaches one of these states $f\in\mathcal{F}$, it receives a lump reward $r(f)$ and transitions to landing mode. The final states denote the desired landing regions for the vehicle, and the corresponding reward captures the scientific interest of the landing zone.
We also model all infeasible states $(x,y,z)\not \in \mathcal{S}$ (i.e., all states not in navigable water or outside the domain of the model) as final states, associated with a negative reward. 

\paragraph{Transitions}
We model the vehicle as a semi-Lagrangian tracer, where the horizontal dynamics are driven by the flow field, and the vertical dynamics are controlled through the variable buoyancy mechanism.
We leverage the flow field model described in Section \ref{sec:cavity-model} to capture the stochastic dynamics that the vehicle may encounter. Specifically, we assume that the model captures the \emph{likely distribution} of the flow field at every point in the state space. Due to uncertainty in the initial and boundary conditions, the model cannot accurately reproduce the flow that will be encountered by the vehicle at a specific time; however, we assume that the empirical temporal distribution of flow velocities encountered over the simulation is representative of the probabilistic distribution of velocities that the vehicle may encounter. Rigorously, define $\vec{v}(x,y,z)$ as a random variable denoting the flow velocity encountered by the vehicle at state $(x,y,z)$, and denote as $\tilde{v}(t,x,y,z)$ the flow velocity predicted by the numerical model described in  Section \ref{sec:cavity-model}. We assume that
\[
P(\vec v(x,y,z) = v) \propto \int_{t_0}^{t_f} 1_{\tilde{v}(t,x,y,z)=v} dt
\]
where ${t_0}$ and ${t_f}$ are the temporal boundaries of the cavity model simulation, and $1_{x}$ is a Boolean function assuming value 1 if $x$ is true and 0 otherwise.

For a given state $(x,y,z)\in \mathcal{S}$, action $\mdpaction\in \mathcal{A}((x,y,z))$, and realization of the velocity field $\vec{v}(x,y,z)$, we model the vehicle transition as
\begin{equation}
s' = (x',y', z') = [x+ \vec{v}_x(x,y,z) \delta, y+ \vec{v}_y(x,y,z) \delta, \mdpaction],
\label{eq:transition-model}
\end{equation}
where $\vec{v}_x$ and $\vec{v}_y$ are the components of the velocity vector $\vec{v}$ along $x$ and $y$ respectively. If $s'\not\in \mathcal{S}$, the vehicle transitions to a final state associated with a negative reward, as discussed above.
Accordingly, the probability of transitioning to state $(x',y',z')\in \mathcal{S}$ from state $(x,y,z)\in \mathcal{S}$ with action $\mdpaction\in \mathcal{A}(s)$ can be computed as
\begin{align}
&P((x',y',z')| (x,y,z), \mdpaction) = & \label{eq:transition-probability}\\
 &\quad 1_{z'=\mdpaction} \cdot \int_{\mathrlap{\zeta: (x,y,\zeta)\in\mathcal{S}}} P(\vec{v}(x,y,z) = [(x'-x) /\delta,  (y'-y)/\delta, \zeta]) d\zeta ,\nonumber
\end{align}
\frmargin{that is, the probability of transitioning from $(x,y,z)$ to $(x',y',z')$ equals the probability of encountering a flow velocity $\vec v$ such that $\vec {v}_x \delta =x'-x$ and $\vec {v}_y \delta =y'-y$ if the commanded depth is $a=z'$, and is zero otherwise.}{Add, polish, etc}
  


\paragraph{Rewards}
Each state-action pair is associated with a reward that captures the energy cost of the action undertaken. The energy cost consists of a constant term  $e_h$ that captures the ``hotel load" required by the vehicle for non-propulsion purposes (e.g., computing and localization), and a variable term $e_b$ that captures the energy used by the buoyancy control mechanism.
The buoyancy control mechanism consumes virtually no energy to descend (as water pressure is used to force oil from the external bladder to the pressure vessel after a valve is opened); in contrast, when the vehicle ascends, a pump works against the water pressure. In this paper, we adopt a simple model where the energy cost is proportional to the desired change in depth with proportional constant $\alpha_b$; the adoption of a more sophisticated energy model is an interesting direction for future research. 
Formally, the reward for the state-action pair $(x,y,z) \in \mathcal{S}, \mdpaction \in\mathcal{A}(s)$ is
\[
r((x,y,z),\mdpaction) = e_h + \alpha_b \cdot \max(\mdpaction - z, 0).
\]

\paragraph{Discussion}
A few comments are in order. 
First, the proposed approach strongly relies on the cavity flow model to capture the probabilistic distribution of the flow dynamics. Accordingly, a representative model that is able to characterize both the likely under-ice flow and its variability is critical to achieve good performance.
Second, the approach does not exploit spatial or temporal correlations in the model: that is, knowledge of the flow encountered by the vehicle at one location is not used to update transition probabilities at other nearby locations, and seasonal effects are averaged out. While this choice helps avoid overfitting the model output, exploiting spatial and temporal correlations in a principled way can help improve performance, and is a highly promising direction for future research.
Third, we use a  simple one-step integration scheme to compute transitions in Equation \eqref{eq:transition-model}. A more refined integration scheme that updates vertical velocities within the time step may offer additional fidelity, and is an interesting direction for future research. 
 Finally, we use a simple model for energy costs; we remark that the proposed modeling approach can accommodate arbitrarily sophisticated energy models with no structural changes.

\subsection{Approximate Dynamic Programming Solution}
\label{sec:navigation:adp}

We are now in a position to solve the continuous MDP through approximate dynamic programming (ADP).
We discretize the state space in a discrete set of states $\tilde S$ forming a uniform lattice. We remark that the ADP discretization needs not correspond to the discretization used in the cavity model.

For each state $\tilde{s} \in \tilde{\mathcal{S}}$, the optimal value of the state (i.e., the optimal discounted expected reward that an agent will obtain when departing from that state) can be computed through the Bellman equation as
\begin{equation}
V^\star(\tilde s) = \max_{a\in\mathcal{A}(\tilde s)} \left( r(\tilde s, a) + \gamma \mathbb{E}_{s'\sim P(s'|\tilde s, a) }\!\left[ V^\star(s') \right] \right),
\label{eq:optimal-value}
\end{equation}
and the optimal action for state $\tilde s$ is
\begin{equation}
a^\star(\tilde s) = \arg\max_{a\in\mathcal{A}(\tilde s)} \left( r(\tilde s, a) + \gamma \mathbb{E}_{s'\sim P(s'|\tilde s, a) }\!\left[ V^\star(s') \right] \right),
\label{eq:optimal-action}
\end{equation}
where $\gamma\in(0,1)$ is the discount factor.

The value of states $s'\not\in\tilde S$ is computed by linearly interpolating the values of states in $\tilde S$. Recall that the states in $\tilde S$ form a regular lattice. Denote as $\tilde{\mathcal{N}}(s')\subset \tilde S$ the states that form the vertices of the lattice cell that contains $ s'$. Then the optimal value of $s'$ is approximated as
\begin{subequations}
\begin{align}
&V^\star(s') = \sum_{\tilde s \in \tilde{\mathcal{N}}(s')  } \lambda_{\tilde s} V^\star(\tilde s), \quad \text{ where}
\label{eq:optimal-value-approx} \\
&\lambda_{\tilde s} \propto \frac{1}{\|s'-\tilde s\|} \quad \forall \tilde s\in \tilde{\mathcal{N}}(s') \quad \text{and} \quad
\sum_{\tilde s \in \tilde{\mathcal{N}}} \lambda_{\tilde s} = 1
\label{eq:optimal-value-coeffs}
\end{align}
\label{eq:optimal-value-approx-and-coeffs}
\end{subequations}

Equations \eqref{eq:optimal-value}-\eqref{eq:optimal-value-approx-and-coeffs} are solved via value iteration, yielding an optimal policy for under-ice guidance that provides an optimal action $a^\star(\tilde s)$ for every state $\tilde s\in\tilde{\mathcal{S}}$. For states not in $\tilde{\mathcal{S}}$, a nearest-neighbor approach is used whereby the policy corresponding to the closest state in $\tilde{\mathcal{S}}$ is used.

\subsection{State uncertainty: a QMDP approach}
\label{sec:navigation:qmdp}
The policy computed in Section \ref{sec:navigation:adp} requires perfect knowledge of the location of the IceNode. In contrast, the location of underwater vehicles typically presents a significant degree of uncertainty. 
IceNodes can estimate their location through acoustic multilateration from moored sound sources placed at the shelf edge; the technique yields uncertainties on the order of $0.5$ km in the radial direction and $D/40$ in the azimuthal direction from the moored buoys, where $D$ is the distance from the buoy \cite{Girton2019-js}.

To address this uncertainty, we propose using the QMDP algorithm \cite{littman1995learning}\frmargin{, which is well-suited for the  embedded, power-constrained IceNode platform due to its modest computational requirements.}{Added}
Intuitively, for a given belief over the vehicle location, QMDP selects the action that yields the best expected value, where the expectation is taken over the states where the vehicle may be.
Rigorously, let the vehicle's belief over its location be the probability distribution $\mathcal{B}$. Let the set of available actions be
\[
\mathcal{A}(\mathcal{B}) = \bigcap_{s\in\mathcal{S}: \mathcal{B}( s)>0} \mathcal{A}(s)
\]
Then, we select the action for belief $\mathcal{B}$ as:
\begin{equation}
a^\star(\mathcal{B}) = \arg \max_{a\in\mathcal{A}(\mathcal{B})}  \mathbb{E}_{s\sim \mathcal{B}}  \left[ r(s, a) + \gamma \mathbb{E}_{s'\sim P(s'| s, a) }\left[ V^\star(s') \right] \right],
\label{eq:qmdp-policy}
\end{equation}
where $V^\star(s')$ is computed according to \eqref{eq:optimal-value}.

We remark that evaluating the optimal policy \eqref{eq:qmdp-policy} requires minimal computational effort, since the optimal state values $V^\star$ can be pre-computed and stored: therefore, the proposed approach is well-suited for on-board guidance of vehicles with highly limited computation resources.

A key limitation of the QMDP formulation is that it assumes that all uncertainty will disappear at the next time step: hence, the approach is unable to perform information-gathering actions (e.g., improving localization by steering towards areas where the flow is well-characterized, and then comparing the actual motion experienced by the vehicle with the model).
An interesting direction for future research will encompass the use of more sophisticated POMDP algorithms such as Monte Carlo Tree Search \cite{Browne2012MCTS} to assess  the effectiveness of such information-gathering actions.

\section{Numerical Experiments}
 \label{sec:experiments}
 
\begin{figure}[h]
 \centering
 \includegraphics[width=.4\textwidth]{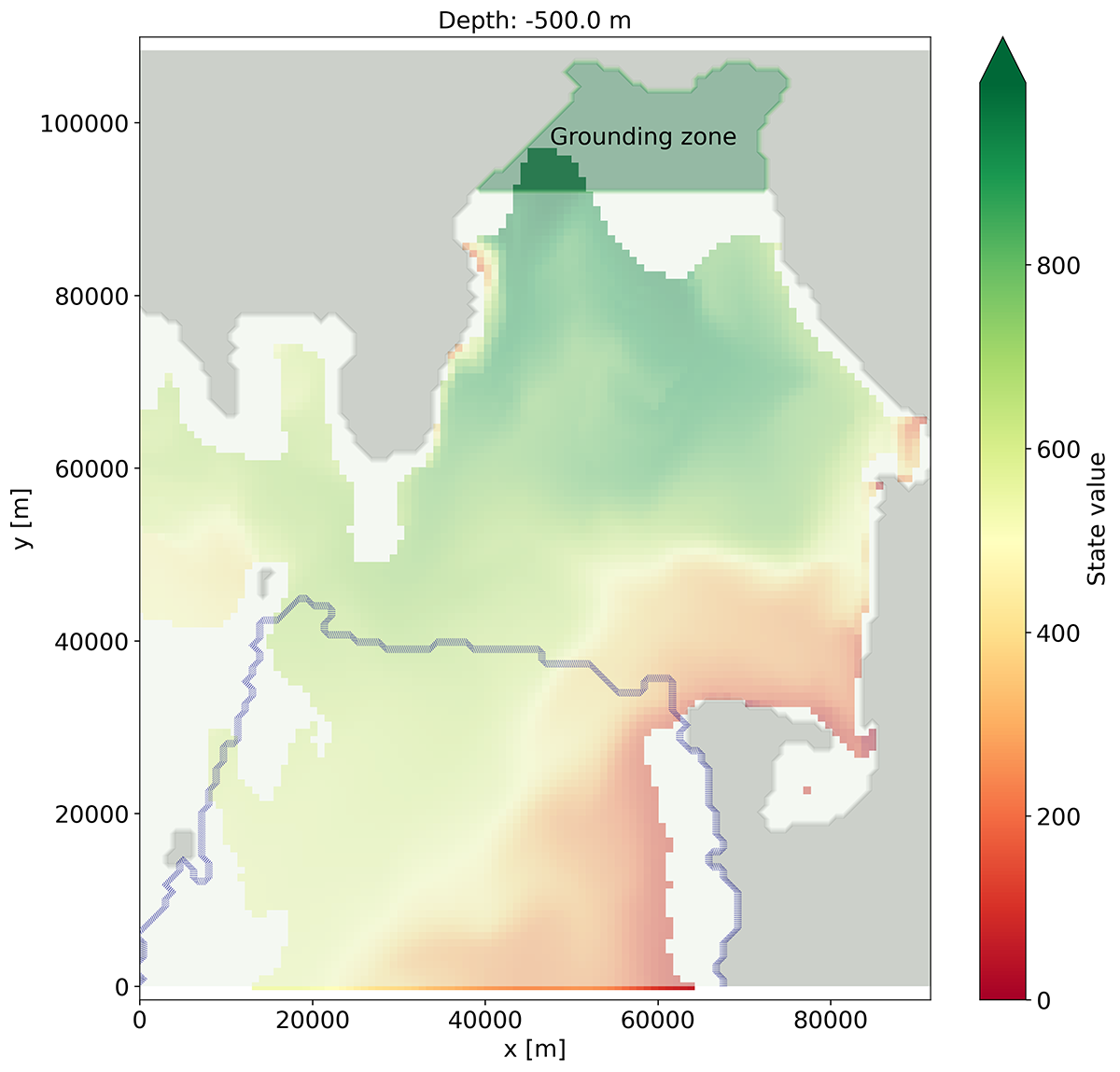}
 \caption{MDP problem state value for $z = - 500$ m. The color of each location denotes the expected discounted reward obtained when following the optimal policy from that location.}
\label{fig:adp:values}
 \end{figure}

\begin{figure}[h]
 \centering
 \includegraphics[width=.4\textwidth]{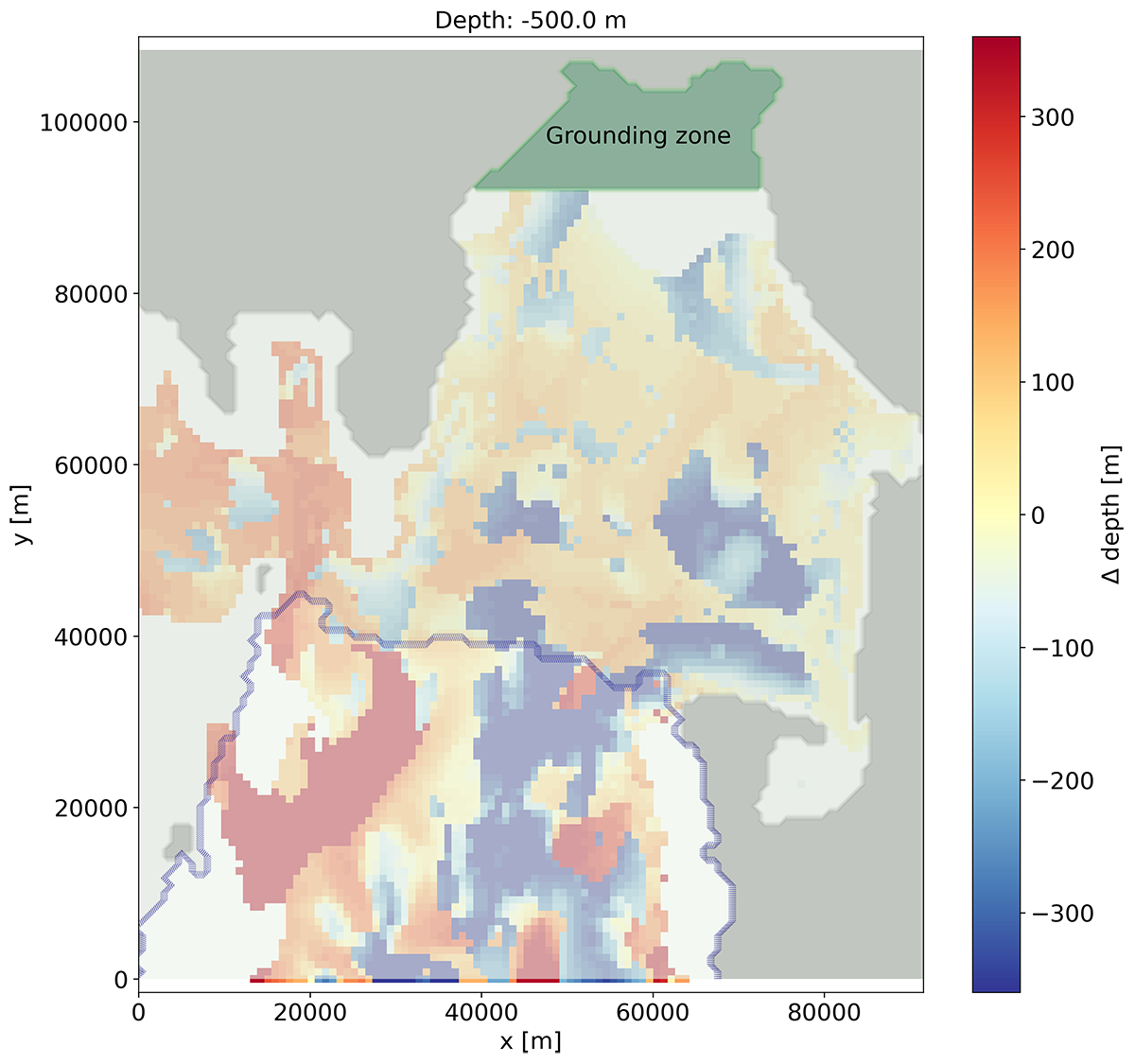}
 \caption{MDP problem policy for $z = - 500$ m. Color denotes the change in depth prescribed by the optimal policy.}
\label{fig:adp:policy}
 \end{figure}

 We characterize the performance of the proposed approach through numerical simulations. 
Due to space constraints, we focus our analysis on the problem of reaching the grounding zone; the dual problem of egress from the grounding zone to open sea will be the subject of future studies.
 We solve the ADP problem \eqref{eq:optimal-value}-\eqref{eq:optimal-value-approx-and-coeffs} on a lattice with a stride of $840 \times 840 \times 25$ m\frmargin{, which we empirically found to provide a good balance between computational and storage cost and policy performance}{polish}. 
To compute the transition probabilities, we use 20\% of all available time steps (i.e., 1752 steps, or one step every five hours), to capture the fact that model knowledge may not perfectly reproduce the actual flow, especially for what concerns short-term and small-scale dynamics.
 The resulting state values and optimal policy are shown in Figures \ref{fig:adp:values} and \ref{fig:adp:policy} for one selected depth. 

     \begin{figure*}[h]
 \centering
 \includegraphics[width=\textwidth]{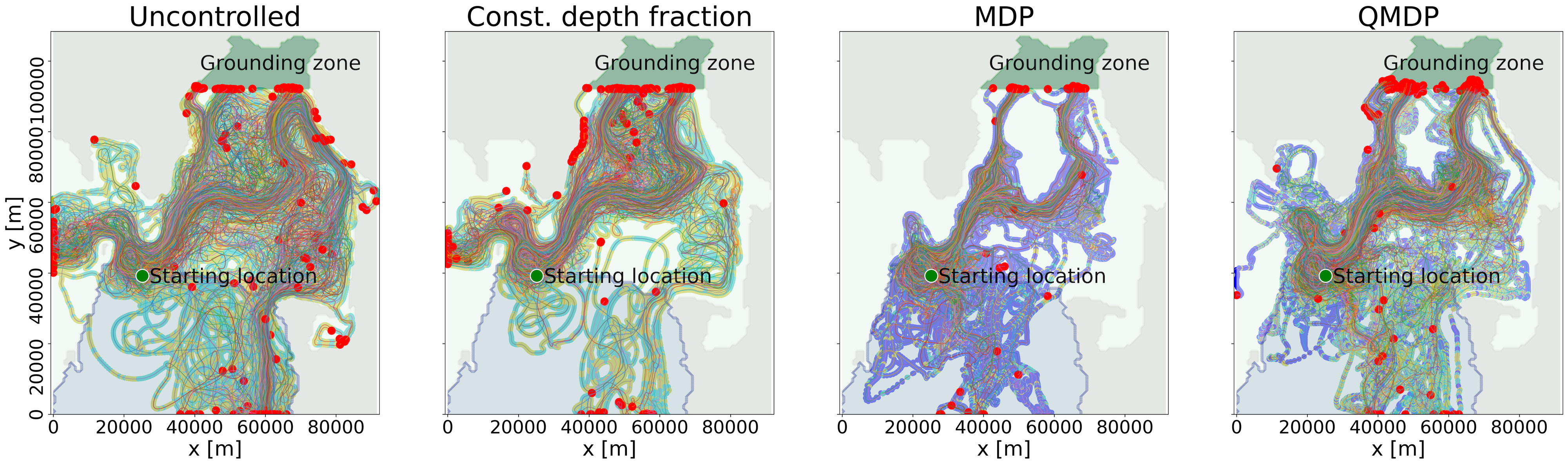}
 \includegraphics[width=\textwidth]{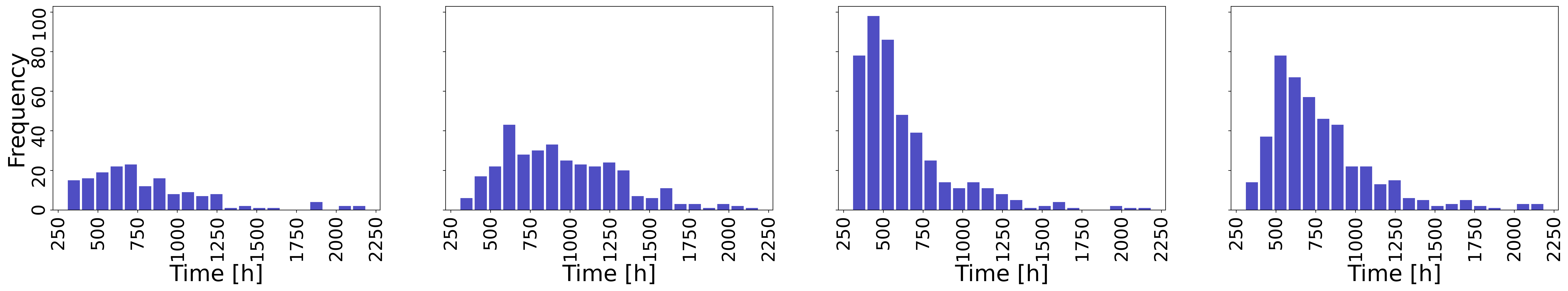}
 \caption{Policy rollouts and time required to reach the landing zone for successful rollouts. For each policy, 500 IceNode trajectories are simulated. The color of the trajectory shows the change in depth, either through a control action or vertical forcing due to current: yellow corresponds to an ascent, blue captures constant-depth drifting, and cyan shows a descent. Red dots show the vehicles' final locations.}
\label{fig:rollouts}
\jvspace{-2em}
 \end{figure*}
 
%
%

  
 We compare the performance of the proposed QMDP policy with three other policies:
 \begin{itemize}
 \item an \emph{uncontrolled} policy where the vehicle drifts in the current with no active buoyancy control;
 \item the state-of-the-art \emph{constant depth fraction} policy implemented by the 2019 APL-UW Dotson Ice Shelf EM-APEX campaign \cite{Girton2019-js}, where the vehicle controls its buoyancy to float at a depth corresponding to 75\% of the cavity depth. In the implementation, we assume that the vehicle has perfect knowledge of its location and the seafloor and basal ice depth, resulting in an \emph{upper bound} on the effectiveness of the  policy.
 \item the \emph{MDP policy} where the vehicle follows Equation \eqref{eq:optimal-action} with perfect knowledge of its location. The MDP policy represents an \emph{upper bound} on the performance of the QMDP policy, and it allows us to quantitatively assess the value of knowledge about the vehicle's position.
 \end{itemize}

For each policy, we perform 500 rollouts. In each rollout, we pick a random initial time and let the cavity flow (and the vehicle position) evolve according to the MITgcm model from that time onwards. The simulation uses \emph{all} available time steps, capturing shorter-term dynamics that are not available to the MDP model. Vehicles that have not reached the grounding zone after three months are assumed to be lost.
All rollouts begin at a manually-selected starting location that mimics state-of-the-art deployment strategies for under-ice vehicles. Specifically, the starting location and depth are selected to be close to the inlet of the cavity, in the region where the most robust inflow current exists, maximizing the likelihood that the vehicle will be dragged deep beneath the shelf.
For the QMDP policy, the vehicle's belief about its location follows a Gaussian distribution with $\sigma_x\!=\!\sigma_y\!=\!1000$ m and  $\sigma_z\!=\!3$ m, which is \frmargin{consistent with the localization performance demonstrated by the EM-APEX campaign \cite{Girton2019-js}.}{Was: preliminary studies show that acoustic multilateration and on-board filtering can yield localization performance well in excess of this threshold on the IceNode platform. }

Results are shown in Figure \ref{fig:rollouts} and in Table \ref{tab:rollouts}.
 
 \begin{table}[h]
  \jvspace{-1em}
\centering
\caption{Performance of underwater guidance policies.} 
\label{tab:rollouts}
\begin{tabular}{r|ccc}
&Reached&\multicolumn{2}{c}{Time to GZ [h]} \\
&  grounding zone & Median & Std. dev\\
\hline
Uncontrolled &33.8\%& 725& 435 \\
Const. depth fraction & 66.6\%& 890 & 383 \\
MDP &95.4\% & 517& 300 \\
\textbf{QMDP} & 88.8\% & 702 & 325  
\end{tabular}
 \jvspace{-.75em}
\end{table}

The proposed QMDP policy is able to deliver close to 90\% of all vehicles to the landing zone - a performance well in excess of the state-of-the-art constant depth fraction policy's, and over 2.5 times as good as the uncontrolled policy. The proposed approach also delivers IceNodes to the grounding zone 26\%, or eight days, faster than the constant depth fraction policy, resulting in increased science returns. Imperfect position knowledge results in an 6.6\% reduction in the success rate of the proposed guidance policy, and a 36\% increase in the median navigation time, compared to the MDP policy; this motivates the study of model-based \emph{localization} techniques to further reduce position uncertainty and approach the performance of the MDP policy.

Figure \ref{fig:rollouts} shows the trajectories produced by the four policies. The MDP policy sharply exploits the structured nature of under-ice currents, and the resulting trajectories are highly clustered around two favorable sets of paths. Remarkably, the trajectories produced by the QMDP policy present a similar qualitative distribution; however, position uncertainty results in several IceNodes being swept out to sea or in side cavities. The constant depth fraction policy is highly effective at delivering vehicles under the ice shelf; however, only a fraction of the vehicles make it to the grounding zone, whereas many more are swept to side cavities or grounded against the sides of the cavity. Finally, despite the selection of a favorable starting location, the uncontrolled policy is only marginally effective at delivering vehicles to the grounding zone, with the majority of IceNodes adrift, lost to side cavities, or swept to sea.
  
  \jvspace{-.3em}
 \section{Conclusions}
 \label{sec:conclusions}
We presented a novel approach for guidance of buoyancy-controlled vehicles under ice shelves in uncertain ocean currents. The proposed technique estimates the probabilistic distribution of ocean currents by leveraging  numerical simulations of the ice cavity flow, and it can cope with realistic uncertainty in the vehicle's localization estimate. Numerical simulations show that the technique significantly outperforms existing under-ice guidance techniques, and holds promise to allow reliable and cost-effective access to ice shelf grounding zones, which hold the key to  better understanding ice shelf melt rates and improving predictions of future sea level rise.

A number of directions for future research are of interest.
First, we plan to further extend the approach to capture the effect of bathymetry uncertainty. Bathymetry uncertainty introduces two sources of error: first, regions that are assumed to be navigable may be occupied by ice or rock, and vice versa; second, uncertainty in the bathymetry profile induces significant uncertainty in the currents, especially near the boundaries. We will \frmargin{quantify}{was: investigate} both effects by leveraging numerical simulations on reduced-resolution cavity models, and  incorporate these sources of uncertainty in the MDP model \frmargin{to mitigate their impact on the policy's performance}{added}.
Second, we will further explore partially observable MDP approaches, and assess whether IceNode's observations can be used to improve the knowledge of its location by exploiting the cavity flow model.
Third, we will consider reinforcement learning approaches where the IceNode's observations are used to improve the flow field model during navigation, leveraging spatial and temporal correlations in the flow field. To support this, we will consider fast online algorithms to re-solve the guidance problem on board the vehicle with minimal energy and time expenditure.
Finally, we plan to validate the approach through field tests in open ocean, first using a virtually injected ice shelf, and later during field deployments beneath real-world ice shelves.

\jvspace{-.3em}
\section*{Acknowledgements}
Part of this work was carried out at the Jet Propulsion Laboratory (JPL), California Institute of Technology, 
 under a contract with the National Aeronautics and Space Administration (80NM0018D0004).
We gratefully acknowledge support from the NASA Cryospheric Sciences Programs. High-end computing resources were provided by the NASA Advanced Supercomputing (NAS) Division of the Ames Research Center. 

\jvspace{-.7em}
 {\small
\bibliographystyle{IEEEtran}
\bibliography{main}
}

\end{document}